\documentclass[runningheads]{llncs}
\usepackage{graphicx}

\usepackage{tikz}
\usepackage{comment} 
\usepackage{amsmath,amssymb} 
\usepackage{color}
\usepackage{multirow}
\usepackage{appendix}

\begin{document}
\pagestyle{headings}
\mainmatter
\def\ECCVSubNumber{2290}  

\title{EPNet: Enhancing Point Features with Image Semantics for 3D Object Detection} 

\titlerunning{EPNet: Enhancing Point Features with Image Semantics}
%

\author{Tengteng Huang\inst{1}$^{\star}$  \and
	Zhe Liu\inst{1}\thanks{equal contribution} \and
	Xiwu Chen\inst{1} \and Xiang Bai\inst{1}\thanks{corresponding author}}

\authorrunning{T. Huang, Z. Liu, et al.}

\institute{Huazhong University of Science and Technology \\
	\email{\{huangtengtng, zheliu1994, xiwuchen, xbai\}@hust.edu.cn}}

\maketitle

\begin{abstract}
In this paper, we aim at addressing two critical issues in the 3D detection task, including the exploitation of multiple sensors~(namely LiDAR point cloud and camera image), as well as the inconsistency between the localization and classification confidence. To this end, we propose a novel fusion module to enhance the point features with semantic image features in a point-wise manner without any image annotations. Besides, a consistency enforcing loss is employed to explicitly encourage the consistency of both the localization and classification confidence. We design an end-to-end learnable framework named EPNet to integrate these two components. Extensive experiments on the KITTI and SUN-RGBD datasets demonstrate the superiority of EPNet over the state-of-the-art methods. 
Codes and models are available at:
\url{https://github.com/happinesslz/EPNet}.
\end{abstract}

\section{Introduction}

\begin{figure}[t]
\begin{center}
  \includegraphics[width=1.0\linewidth]{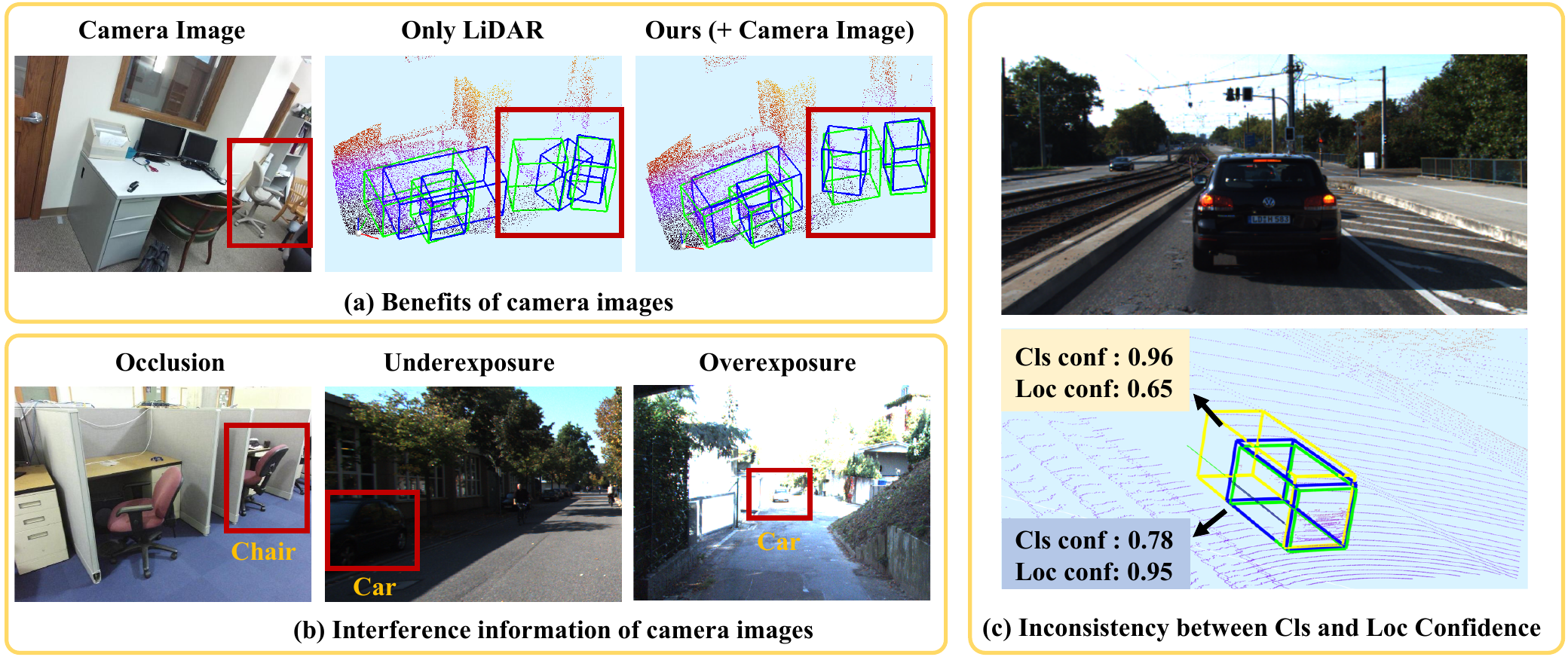}
\end{center}
\setlength{\abovecaptionskip}{2pt}
\caption{Illustration of (a)~the benefit and (b)~potential interference information of the camera image. (c) demonstrates the inconsistency of the classification confidence and localization confidence. Green box denotes the ground truth. Blue and yellow boxes are predicted bounding boxes.} 
\label{fig:introduction}
\end{figure}

The last decade has witnessed significant progress in the 3D object detection task via different types of sensors, such as monocular images~\cite{chen2016monocular,xu2018multi}, stereo cameras~\cite{chen20173d}, and LiDAR point clouds~\cite{zhou2018voxelnet,luo2018fast,yang2018pixor}. Camera images usually contain plenty of semantic features~(\textit{e.g.}, color, texture) while suffering from the lack of depth information. LiDAR points provide depth and geometric structure information, which are quite helpful for understanding 3D scenes. However, LiDAR points are usually sparse, unordered, and unevenly distributed. Fig.~\ref{fig:introduction}(a) illustrates a typical example of leveraging the camera image to improve the 3D detection task. It is challenging to distinguish between the closely packed white and yellow chairs by only the LiDAR point cloud due to their similar geometric structure, resulting in chaotically distributed bounding boxes. In this case, utilizing the color information is crucial to locate them precisely. This motivates us to design an effective module to fuse different sensors for a more accurate 3D object detector.

However, fusing the representations of LiDAR and camera image is a non-trivial task for two reasons. On the one hand, they possess highly different data characteristics. On the other hand, the camera image is sensitive to illumination, occlusion, \textit{etc.}~(see Fig.~\ref{fig:introduction}(b)), and thus may introduce interfering information that is harmful to the 3D object detection task. Previous works usually fuse these two sensors with the aid of image annotations~(namely 2D bounding boxes). According to different ways of utilizing the sensors, we summarize previous works into two main categories, including 1) cascading approaches using different sensors in different stages~\cite{qi2018frustum,xu2017pointfusion,zhao20193d}, and 2) fusion methods that jointly reason over multi-sensor inputs~\cite{liang2019multi,Liang2018ECCV}. Although effective, these methods have several limitations. Cascading approaches cannot leverage the complementarity among different sensors, and their performance is bounded by each stage. Fusion methods~\cite{liang2019multi,Liang2018ECCV} need to generate BEV data through perspective projection and voxelization, leading to information loss inevitably. Besides, they can only approximately establish a relatively coarse correspondence between the voxel features and semantic image features. We propose a LiDAR-guided Image Fusion~(LI-Fusion) module to address both the two issues mentioned above. LI-Fusion module establishes the correspondence between raw point cloud data and the camera image in a point-wise manner, and adaptively estimate the importance of the image semantic features. In this way, useful image features are utilized to enhance the point features while interfering image features are suppressed. Comparing with previous method, our solution possesses four main advantages, including 1) achieving fine-grained point-wise correspondence between LiDAR and camera image data through a simpler pipeline without complicated procedure for BEV data generation; 2) keeping the original geometric structure without information loss; 3) addressing the issue of the interference information that may be brought by the camera image; 4) free of image annotations, namely 2D bounding box annotations, as opposed to previous works~\cite{qi2018frustum,Liang2018ECCV}.


Besides multi-sensor fusion, we observe the issue of the inconsistency between the classification confidence and localization confidence, which represent whether an object exists in a bounding box and how much overlap it shares with the ground truth. As shown in Fig.~\ref{fig:introduction}(c), the bounding box with higher classification confidence possesses lower localization confidence instead. This inconsistency will lead to degraded detection performance since the Non-Maximum Suppression~(NMS) procedure automatically filters out boxes with large overlaps but low classification confidence. However, this problem is rarely discussed in the 3D detection task. Jiang \textit{et al.}~\cite{jiang2018acquisition} attempt to alleviate this problem by improving the NMS procedure. They introduce a new branch to predict the localization confidence and replace the threshold for the NMS process as a multiplication of both the classification and localization confidences. Though effective to some extent, there is no explicit constraint to force the consistency of these two confidences. Different from ~\cite{jiang2018acquisition}, we present a consistency enforcing loss~(CE loss) to guarantee the consistency of these two confidences explicitly. With its aid, boxes with high classification confidence are encouraged to possess large overlaps with the ground truth, and vice versa. This approach owns two advantages. First, our solution is easy to implement without any modifications to the architecture of the detection network. Second, our solution is entirely free of learnable parameters and extra inference time overhead.

Our key contributions are as follows:
\begin{enumerate}
\item Our LI-Fusion module operates on LiDAR point and camera image directly and effectively enhances the point features with corresponding semantic image features in a point-wise manner without image annotations.
\item We propose a CE loss to encourage the consistency between the classification and localization confidence, leading to more accurate detection results.  
\item We integrate the LI-Fusion module and CE loss into a new framework named EPNet, which achieves state-of-the-art results on two common 3D object detection benchmark datasets, \textit{i.e.}, the KITTI dataset~\cite{geigerwe} and SUN-RGBD dataset~\cite{song2015sun}.

\end{enumerate}

\section{Related Work}

\noindent\textbf{3D object detection based on camera images.}~Recent 3D object detection methods pay much attention to camera images, such as monocular~\cite{ma2019accurate,qin2019monogrnet,ku2019monopsr,li2019gs3d,liu2019deep} and stereo images~\cite{licvpr2019,wangcvpr2019}. Chen \textit{et al.}~\cite{chen2016monocular} obtain 2D bounding boxes with a CNN-based object detector and infer their corresponding 3D bounding boxes with semantic, context, and shape information. Mousavian \textit{et al.}~\cite{mousavian20173d} estimate localization and orientation from 2D bounding boxes of objects by exploiting the constraint of projective geometry. However, methods based on the camera image have difficulty in generating accurate 3D bounding boxes due to the lack of depth information.

\noindent\textbf{3D object detection based on LiDAR.}~Many LiDAR-based methods~\cite{yang2018pixor,lasernet,std2019yang} are proposed in recent years. VoxelNet~\cite{zhou2018voxelnet} divides a point cloud into voxels and employs stacked voxel feature encoding layers to extract voxel features. SECOND~\cite{yan2018second} introduces a sparse convolution operation to improve the computational efficiency of ~\cite{zhou2018voxelnet}. PointPillars~\cite{lang2019pointpillars} converts the point cloud to a pseudo-image and gets rid of time-consuming 3D convolution operations. PointRCNN~\cite{shi2019pointrcnn} is a pioneering two-stage detector, which consists of a region proposal network~(RPN) and a refinement network. The RPN network predicts the foreground points and outputs coarse bounding boxes which are then refined by the refinement network. However, LiDAR data is usually extremely sparse, posing a challenge for accurate localization.

\noindent\textbf{3D object detection based on multiple sensors.} Recently, much progress has been made in exploiting multiple sensors, such as camera image and LiDAR. Qi \textit{et al.}~\cite{qi2018frustum} propose a cascading approach F-PointNet, which first produces 2D proposals from camera images and then generates corresponding 3D boxes based on LiDAR point clouds. However, cascading methods need extra 2D annotations, and their performance is bounded by the 2D detector. Many methods attempt to reason over camera images and BEV jointly. MV3D~\cite{chen2017multi} and AVOD~\cite{ku2018joint} refine the detection box by fusing BEV and camera feature maps for each ROI region. ConFuse~\cite{Liang2018ECCV} proposes a novel continuous fusion layer that achieves the voxel-wise alignment between BEV and image feature maps. Different from previous works, our LI-Fusion module operates on LiDAR data directly and establishes a finer point-wise correspondence between the LiDAR and camera image features.

\section{Method}

Exploiting the complementary information of multiple sensors is important for accurate 3D object detection. Besides, it is also valuable to resolve the performance bottleneck caused by the inconsistency between the localization and classification confidence.

In this paper, we propose a new framework named EPNet to improve the 3D detection performance from these two aspects. EPNet consists of a two-stream RPN for proposal generation and a refinement network for bounding box refining, which can be trained end-to-end. The two-stream RPN effectively combines the LiDAR point feature and semantic image feature via the proposed LI-Fusion module. Besides, we provide a consistency enforcing loss~(CE loss) to improve the consistency between the classification and localization confidence. In the following, we present the details of our two-steam RPN and refinement network in subsection~\ref{sec:image fusion} and subsection~\ref{sec:refinement}, respectively. Then we elaborate our CE loss and the overall loss function in subsection~\ref{sec:overall loss function}.


\subsection{Two-stream RPN}
\label{sec:image fusion}

\begin{figure*}[t]
\begin{center}
  \includegraphics[width=0.95\linewidth]{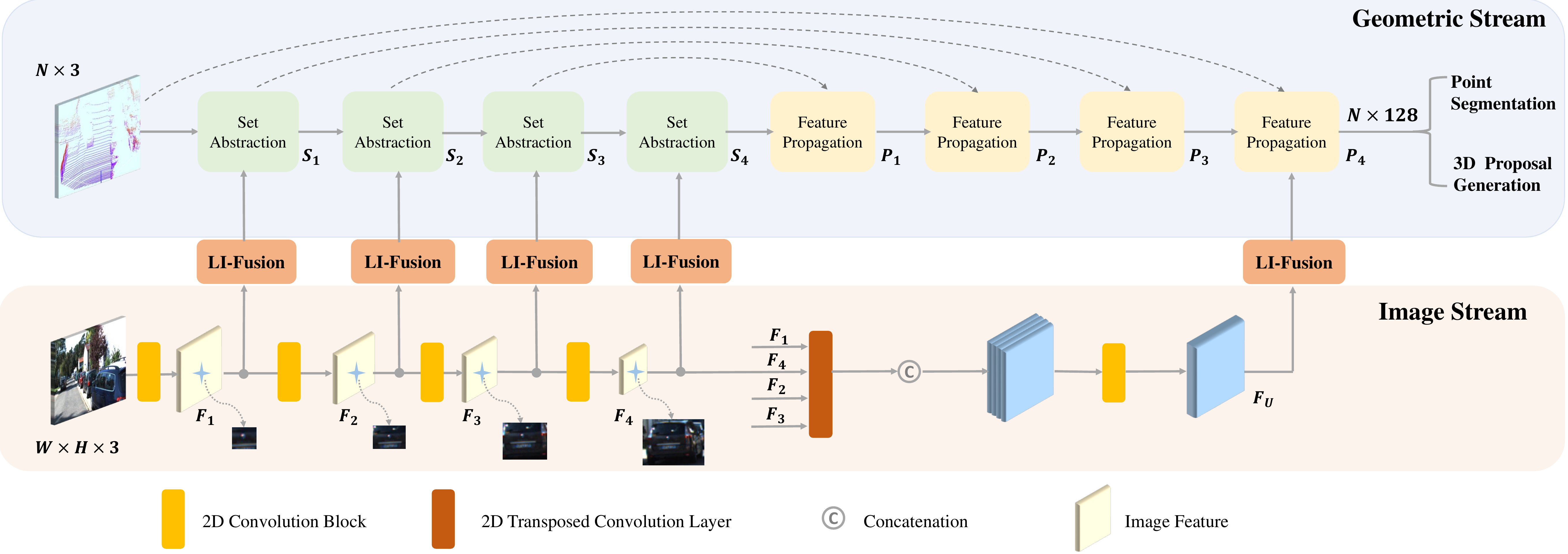}
\end{center}
\setlength{\abovecaptionskip}{2pt}
\caption{Illustration of the architecture of the two-stream RPN which is composed of a geometric stream and an image stream. We employ several LI-Fusion modules to enhance the LiDAR point features with corresponding semantic image features in multiple scales. $N$ represents the number of LiDAR points. $H$ and $W$ denote the height and width of the input camera image, respectively. }
\label{fig:two-stream RPN}
\end{figure*}

Our two-stream RPN is composed of a geometric stream and an image stream. As shown in Fig.~\ref{fig:two-stream RPN}, the geometric stream and the image stream produce the point features and semantic image features, respectively. We employ multiple LI-Fusion modules to enhance the point features with corresponding semantic image features in different scales, leading to more discriminative feature representations.

\noindent\textbf{Image Stream.}~The image stream takes camera images as input and extracts the semantic image information with a set of convolution operations. We adopt an especially simple architecture composed of four light-weighted convolutional blocks. Each convolutional block consists of two $3\times3$ convolution layers followed by a batch normalization layer~\cite{ioffe2015batch} and a ReLU activation function. We set the second convolution layer in each block with stride 2 to enlarge the receptive field and save GPU memory. $F_i$~($i$=1,2,3,4) denotes the outputs of these four convolutional blocks. As illustrated in Fig.~\ref{fig:two-stream RPN}, $F_i$ provides sufficient semantic image information to enrich the LiDAR point features in different scales. We further employ four parallel transposed convolution layers with different strides to recover the image resolution, leading to feature maps with the same size as the original image. We combine them in a concatenation manner and obtain a more representative feature map $F_U$ containing rich semantic image information with different receptive fields. As is shown later, the feature map $F_U$ is also employed to enhance the LiDAR point features to generate more accurate proposals.

\noindent\textbf{Geometric Stream.}~The geometric stream takes LiDAR point cloud as input and generates the 3D proposals. The geometric stream comprises four paired Set Abstraction~(SA)~\cite{qi2017pointnet++} and Feature Propogation~(FP)~\cite{qi2017pointnet++} layers for feature extraction. For the convenience of description, the outputs of SA and FP layers are denoted as $S_i$ and $P_i$~($i$=1,2,3,4), respectively. As shown in Fig.~\ref{fig:two-stream RPN}, we combine the point features $S_i$ with the semantic image features $F_i$ with the aid of our LI-Fusion module. Besides, The point feature $P_4$ is further enriched by the multi-scale image feature $F_U$ to obtain a compact and discriminative feature representation, which is then fed to the detection heads for foreground point segmentation and 3D proposal generation.

\begin{figure}[t]
\begin{center}
  \includegraphics[width=1.0\linewidth]{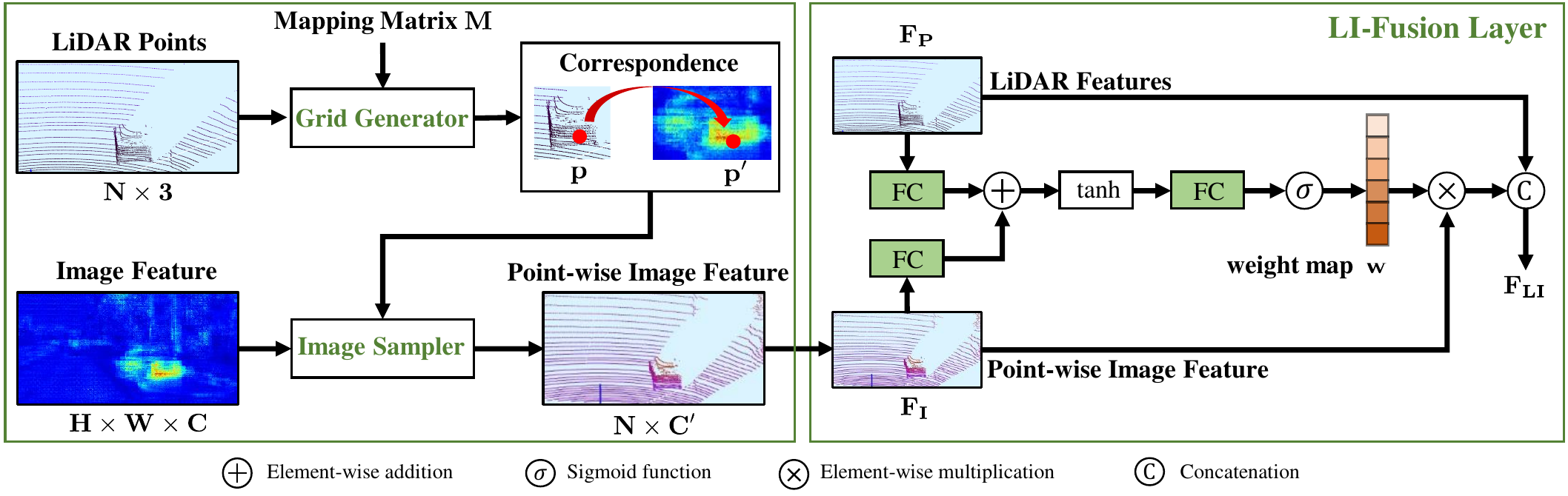}
\end{center}
\setlength{\abovecaptionskip}{2pt}
\caption{Illustration of the LI-Fusion module, which consists of a grid generator, an image sampler, and a LI-Fusion layer.}
\label{fig:LI-Fusion}
\end{figure}

\noindent\textbf{LI-Fusion Module.}~The LiDAR-guided image fusion module consists of a grid generator, an image sampler, and a LI-Fusion layer. As illustrated in Fig.~\ref{fig:LI-Fusion}, the LI-Fusion module involves two parts, \textit{i.e.}, point-wise correspondence generation and LiDAR-guided fusion. Concretely, we project the LiDAR points onto the camera image and denote the mapping matrix as $M$. The grid generator takes a LiDAR point cloud and a mapping matrix $M$ as inputs, and outputs the point-wise correspondence between the LiDAR points and the camera image under different resolutions. In more detail, for a particular point $p(x,y,z)$ in the point cloud, we can get its corresponding position $p'(x', y')$ in the camera image, which can be written as:
\begin{equation}
\label{1}
p' = M \times p ,
\end{equation}
where $M$ is of size $3\times 4$. Note that we convert $p'$ and $p$ into 3-dimensional and 4-dimensional vector in homogeneous coordinates in the projection process formula (\ref{1}).

After establishing the correspondence, we propose to use an image sampler to get the semantic feature representation for each point. Specifically, our image sampler takes the sampling position $p'$ and the image feature map $F$ as inputs to produce a point-wise image feature representation $V$ for each sampling position. Considering that the sampling position may fall between adjacent pixels, we use bilinear interpolation to get the image feature at the continuous coordinates, which can be formularized as follows:

\begin{equation}
V^{(p)}=\mathcal{K}(F^{(\mathcal{N}(p'))}),
\end{equation}
where $V^{(p)}$ is the corresponding image feature for point $p$, $\mathcal{K}$ denotes the bilinear interpolation function, and $F^{(\mathcal{N}(p'))}$ represents the image features of the neighboring pixels for the sampling position $p'$.

Fusing the LiDAR feature and the point-wise image feature is non-trivial since the camera image is challenged by many factors, including illumination, occlusion, etc. In these cases, the point-wise image feature will introduce interfering information. To address this issue, we adopt a LiDAR-guided fusion layer, which utilizes the LiDAR feature to adaptively estimate the importance of the image feature in a point-wise manner. As illustrated in Fig.~\ref{fig:LI-Fusion},  we first feed the LiDAR feature $F_P$ and the point-wise feature $F_I$ into a fully connected layer and map them into the same channel. Then we add them together to form a compact feature representation, which is then compressed into a weight map $\mathbf{w}$ with a single channel through another fully connected layer. We use a sigmoid activation function to normalize the weight map $\mathbf{w}$ into the range of [0, 1].



\begin{equation}
    \mathbf{w} = \sigma (\mathcal{W}\tanh(\mathcal{U}F_P + \mathcal{V}F_I))
\end{equation}
where $\mathcal{W}$, $\mathcal{U}$, $\mathcal{V}$ denote the learnable weight matrices in our LI-Fusion layer. $\sigma$ represents the sigmoid activation function.

After obtaining the weight map $\mathbf{w}$, we combine the LiDAR feature $F_P$ and the semantic image feature $F_I$ in a concatenation manner, which can be formularized as follows:
\begin{equation}
    F_{LI} = F_P \ || \ \mathbf{w} F_I
\end{equation}

\subsection{Refinement Network}
\label{sec:refinement}
We employ the NMS procedure to keep the high-quality proposals and feed them into the refinement network. For each input proposal, we generate its feature descriptor by randomly selecting 512 points in the corresponding bounding box on top of the last SA layer of our two-stream RPN. For those proposals with less than 512 points, we simply pad the descriptor with zeros. The refinement network consists of three SA layers to extract a compact global descriptor, and two subnetworks with two cascaded $1\times 1$ convolution layers for the classification and regression, respectively. 

\subsection{Consistency Enforcing Loss}
\label{sec:intertwined 3d iou loss}
Common 3D object detectors usually generate much more bounding boxes than the number of the real objects in the scene. It poses a great challenge of how to select the high-quality bounding boxes. NMS attempts to filter unsatisfying bounding boxes according to their classification confidence. In this case, it is assumed that the classification confidence can serve as an agent for the real IoU between the bounding and the ground truth, \textit{i.e.}, the localization confidence. However, the classification confidence and the localization confidence is often inconsistent, leading to sub-optimal performance.

This motivates us to introduce a consistency enforcing loss to ensure the consistency between the localization and classification confidence so that boxes with high localization confidence possess high classification confidence, and vice versa. The consistency enforcing loss can be written as follows:


\begin{equation}
L_{ce} = -log(c \times \frac{Area(D \cap G)}{Area(D \cup G)})
\end{equation}

where $D$ and $G$ represents the predicted bounding box and the ground truth. $c$ denotes the classification confidence for $D$. Towards optimizing this loss function, the classification confidence and localization confidence~(\textit{i.e.}, the IoU) are encouraged to be as high as possible jointly. Hence, boxes with large overlaps will possess high classification possibilities and be kept in the NMS procedure. 

\noindent\textbf{Relation to IoU loss.} Our CE loss is similar to the IoU loss~\cite{yu2016unitbox} in the formula, but completely different in the motivation and the function. The IoU loss attempts to generate more precise regression through optimizing the IoU metric, while CE loss aims at ensuring the consistency between the localization  and classification confidence to assist the NMS procedure to keep more accurate bounding boxes. Although with a simple formula, quantitative results and analyses in Sec.~\ref{sec:ablation study} demonstrates the effectiveness of our CE loss in ensuring the consistency and improving the 3D detection performance.

\subsection{Overall Loss Function}
\label{sec:overall loss function}

We utilize a multi-task loss function for jointly optimizing the two-stream RPN and the refinement network. The total loss can be formulated as:
\begin{equation}
L_{total} = L_{rpn} +  L_{rcnn},
\end{equation}
where $L_{rpn}$ and $L_{rcnn}$ denote the training objective for the two-stream RPN and the refinement network, both of which adopt a similar optimizing goal, including a classification loss, a regression loss and a CE loss. We adopt the focal loss~\cite{lin2017focal} as our classification loss to balance the positive and negative samples with the setting of $\alpha=0.25$ and $\gamma=2.0$. For a bounding box, the network needs to regress its center point~$(x, y, z)$, size~$(l, h, w)$, and orientation $\theta$.

Since the range of the Y-axis~(the vertical axis) is relatively small, we directly calculate its offset to the ground truth with a smooth L1 loss~\cite{girshick2015fast}. Similarly, the size of the bounding box~$(h, w,l)$ is also optimized with a smooth L1 loss. As for the X-axis, the Z-axis and the orientation $\theta$, we adopt a bin-based regression loss~\cite{shi2019pointrcnn,qi2018frustum}. For each foreground point, we split its neighboring area into several bins. The bin-based loss first predicts which bin
$b_u$ the center point falls in, and then regress the residual offset $r_u$ within the bin. We formulate the loss functions as follows:
\begin{equation}
L_{rpn}= L_{cls} + L_{reg} + \lambda L_{cf}
\end{equation}
\begin{equation}
    L_{cls} = -\alpha(1-c_t)^\gamma \log c_t
\end{equation}
\begin{equation}
    L_{reg} = \sum_{u \in {x,z, \theta}}E(b_u, \hat{b_u}) + \sum_{u \in {x,y,z,h,w,l,\theta}} S(r_u, \hat{r_u})
\end{equation}
where $E$ and $S$ denote the cross entropy loss and the smooth L1 loss, respectively. $c_t$ is the probability of the point in consideration belong to the ground truth category.  $\hat{b_u}$ and $\hat{r_u}$ denote the ground truth of the bins and the residual offsets. 

\section{Experiments}

We evaluate our method on two common 3D object detection datasets, including the KITTI dataset~\cite{geigerwe} and the SUN-RGBD dataset~\cite{song2015sun}. KITTI is an outdoor dataset, while SUN-RGBD focuses on the indoor scenes. In the following, we first present a brief introduction to these datasets in subsection~\ref{sec:datasets and metric}. Then we provide the implementation details in subsection~\ref{sec:implementation details}. Comprehensive analyses of the LI-Fusion module and the CE loss are elaborated in subsection~\ref{sec:ablation study}.
Finally, we exhibit the comparisons with state-of-the-art methods on the KITTI dataset and the SUN-RGBD dataset in subsection~\ref{sec:experiments on kitti} and subsection~\ref{sec:experiments on sun-rgbd}, respectively.  

\subsection{Datasets and Evaluation Metric}
\label{sec:datasets and metric}

\noindent\textbf{KITTI Dataset} is a standard  benchmark dataset for autonomous driving, which consists of 7,481 training frames and 7,518 testing frames. Following the same dataset split protocol as ~\cite{qi2018frustum,shi2019pointrcnn}, the 7,481 frames are further split into 3,712 frames for training and 3,769 frames for validation. In our experiments, we provide the results on both the validation and the testing set for all the three difficulty levels, \textit{i.e.}, Easy, Moderate, and Hard. Objects are classified into different difficulty levels according to the size, occlusion, and truncation.

\noindent\textbf{SUN-RGBD Dataset} is an indoor benchmark dataset for 3D object detection. The dataset is composed of 10,335 images with 700 annotated object categories, including 5,285 images for training and 5,050 images for testing. We report results on the testing set for ten main object categories following previous works~\cite{xu2017pointfusion,qi2018frustum} since objects of these categories are relatively large.
  
\noindent\textbf{Metrics.} We adopt the Average Precision~(AP) as the metric following the official evaluation protocol of the KITTI dataset and the SUN-RGBD dataset. Recently, the KITTI dataset applies a new evaluation protocol~\cite{simonelli2019disentangling} which uses 40 recall positions instead of the 11 recall positions as before. Thus it is a fairer evaluation protocol. We compare our methods with state-of-the-art methods under this new evaluation protocol.

\subsection{Implementation Details}
\label{sec:implementation details}
\noindent\textbf{Network Settings.}
 The two-stream RPN takes both the LiDAR point cloud and the camera image as inputs. For each 3D scene, the range of LiDAR point cloud is [-40, 40], [-1, 3], [0, 70.4] meters along the X~(right), Y~(down), Z~(forward) axis in camera coordinate, respectively. And the orientation of $\theta$ is in the range of [-$\pi$, $\pi$]. We subsample 16,384 points from the raw LiDAR point cloud as the input for the geometric stream, which is same with PointRCNN~\cite{shi2019pointrcnn}. And the image stream takes images with a resolution of $1280 \times 384$ as input. We employ four set abstraction layers to subsample the input LiDAR point cloud with the size of 4096, 1024, 256, and 64, respectively. Four feature propagation layers are used to recover the size of the point cloud for the foreground segmentation and 3D proposal generation. Similarly, we use four convolution block with stride 2 to downsample the input image. Besides, we employ four parallel transposed convolution with stride 2, 4, 8, 16 to recover the resolution from feature maps in different scales. In the NMS process, we select the top 8000 boxes generated by the two-stream RPN according to their classification confidence. After that, we filter redundant boxes with the NMS threshold of 0.8 and obtain 64 positive candidate boxes which will be refined by the refinement network. For both datasets, we utilize similar architecture design for the two-stream RPN as discussed above.

\noindent\textbf{The Training Scheme.} Our two-stream RPN and refinement network are end-to-end trainable. In the training phase, the regression loss $L_{reg}$ and the CE loss are only applied to positive proposals, \textit{i.e.}, proposals generated by foreground points for the RPN stage, and proposals sharing IoU larger than 0.55 with the ground truth for RCNN stage.

\noindent\textbf{Parameter Optimization.}~ The Adaptive Moment Estimation~(Adam)~\cite{kingma2014adam} is adopted  to optimize our network. The initial learning rate, weight decay, and momentum factor are set to 0.002, 0.001, and 0.9, respectively. We train the model for around 50 epochs on four Titan XP GPUs with a batch size of 12 in an end-to-end manner. The balancing weights $\lambda$ in the loss function are set to 5.

\noindent\textbf{Data Augmentation.}~Three common data augmentation strategies are adopted to prevent over-fitting, including rotation, flipping, and scale transformations. First, we randomly rotate the point cloud along the vertical axis within the range of $[-\pi/18,\pi/18]$. Then, the point cloud is randomly flipped along the forward axis. Besides, each ground truth box is randomly scaled following the uniform distribution of $[0.95,1.05]$. Many LiDAR-based methods sample ground truth boxes from the whole dataset and place them into the raw 3D frames to simulate real scenes with crowded objects following ~\cite{zhou2018voxelnet,yan2018second}. Although effective, this data augmentation needs the prior information of road plane which is usually difficult to acquire for kinds of real scenes. Hence, we do not utilize this augmentation mechanism in our framework for the applicability and generality.


\subsection{Ablation Study}
\label{sec:ablation study}
We conduct extensive experiments on the KITTI validation dataset to evaluate the effectiveness of our LI-Fusion module and CE loss.

\noindent\textbf{Analysis of the fusion architecture.}~We remove all the LI-Fusion modules to verify the effectiveness of our LI-Fusion module.  As is shown in Table~\ref{tab:ablation study}, adding LI-Fusion module yields an improvement of 1.73\% in terms of 3D mAP, demonstrating its effectiveness in combining the point features and semantic image features. We further present comparisons with two alternative fusion solutions in Table~\ref{tab:ablation on different fusion mechanism}. One alternative is simple concatenation~(SC). We modify the input of the geometric stream as the combination of the raw camera image and LiDAR point cloud instead of their feature representations. Concretely, we append the RGB channels of camera images to the spatial coordinate channels of LiDAR point cloud in a concatenation fashion. It should be noted that no image stream is employed for SC. The other alternative is the single scale~(SS) fusion, which shares a similar architecture as our two-stream RPN. The difference is that we remove all the LI-Fusion modules in the set abstraction layers and only keep the LI-Fusion module in the last feature propagation layer~(see Fig.~\ref{fig:two-stream RPN}). As shown in Table~\ref{tab:ablation on different fusion mechanism}, SC yields a decreasement of 3D mAP 0.28\% over the baseline, indicating that simple combination in the input level cannot provide sufficient guidance information. Besides, our method outperforms SS by 3D mAP 1.31\%. It suggests the effectiveness of applying the LI-Fusion modules in multiple scales.

\begin{minipage}{0.94\textwidth}
\hfill
\begin{minipage}{0.47\textwidth}
\makeatletter\def\@captype{table}\makeatother
\setlength{\belowcaptionskip}{4pt}%
\caption{Ablation experiments on the KITTI val dataset. }
\label{tab:ablation study}
\scriptsize
\resizebox{\columnwidth}{!}{
\begin{tabular}{|c|c|c|c|c|c|c|}
\hline
LI-Fusion &CE &Easy &Moderate &Hard &3D mAP &Gain \\\hline\hline
$\times$  &$\times$ &86.34  &77.52 &75.96  &79.94   &-\\ 
\checkmark  &$\times$ &89.44  &78.84 &76.73  &81.67 &\textcolor{blue}{$\uparrow$ 1.73}\\
$\times$ &\checkmark &90.87 &81.15 &79.59   &83.87 &\textcolor{blue}{$\uparrow$ 3.93}  \\
\checkmark &\checkmark &\textbf{92.28} &\textbf{82.59} &\textbf{80.14} &\textbf{85.00} &\textcolor{blue}{$\uparrow$ 5.06}    \\
\hline
\end{tabular}
}
\end{minipage}
\hfill
\begin{minipage}{0.47\textwidth}
\centering
\makeatletter\def\@captype{table}\makeatother
\setlength{\belowcaptionskip}{5pt}%
\caption{Analysis of different fusion mechanism on the KITTI val dataset.}
\label{tab:ablation on different fusion mechanism}
\scriptsize
\resizebox{\columnwidth}{!}{
\begin{tabular}{|c|c|c|c|c|c|c|c|}
\hline
SC & SS & Ours &Easy &Moderate &Hard &3D mAP &Gain\\
\hline\hline
$\times$ &$\times$ &$\times$   &86.34 &77.52 &75.96  &79.94   &- \\
\checkmark &$\times$ &$\times$ &85.97 &77.37 &75.65  &79.66   &\textcolor{red}{$\downarrow$ 0.28} \\
$\times$ &\checkmark  &$\times$ &87.46 &78.27 &75.35 &80.36   &\textcolor{blue}{$\uparrow$ 0.42} \\
$\times$ &$\times$ &\checkmark &\textbf{89.44} &\textbf{78.84} &\textbf{76.73}  &\textbf{81.67} &\textcolor{blue}{ $\uparrow$ 1.73} \\
\hline
\end{tabular}
}
\end{minipage}
\end{minipage}

\begin{figure}[t]
\begin{center}
  \includegraphics[width=0.9\linewidth]{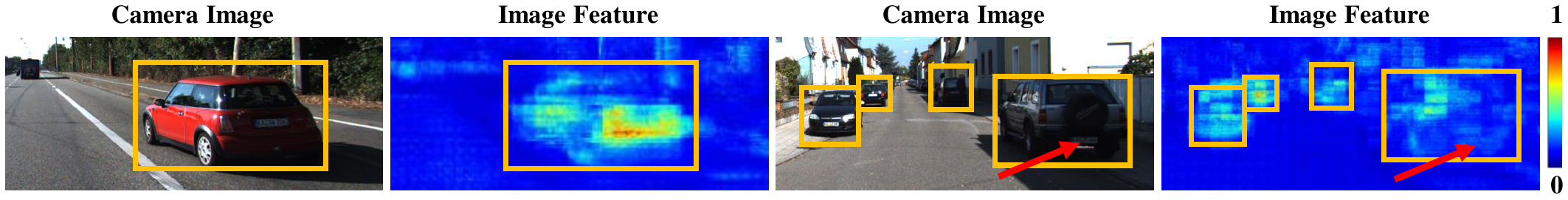}
\end{center}
\setlength{\abovecaptionskip}{2pt}
\caption{Visualization of the learned semantic image feature. The image stream mainly focuses on the foreground objects~(cars). The red arrow marks the region under bad illumination, which show a distinct feature representation to its neighboring region. }
\label{fig:visualized image feature}
\end{figure}

\noindent\textbf{Visualization of learned semantic image features.}~It should be noted that we do not add explicit supervision information~(\textit{e.g.}, annotations of 2D detection boxes) to the image stream of our two-stream RPN. The image stream is optimized together with the geometric stream with the supervision information of 3D boxes from the end of the two-stream RPN. Considering the distinct data characteristics of the camera image and LiDAR point cloud, we visualize the semantic image features to figure out what the image stream learns, as presented in Fig.~\ref{fig:visualized image feature}. Although no explicit supervision is applied, surprisingly, the image stream learns well to differentiate the foreground objects from the background and extracts rich semantic features from camera images, demonstrating that the LI-Fusion module accurately establishes the correspondence between LiDAR point cloud and camera image, thus can provide the complementary semantic image information to the point features. It is also worth noting that the image stream mainly focuses on the representative region of the foreground objects and that the region under poor illumination demonstrates very distinct features to neighboring region, as marked by the red arrow. It indicates that it is necessary to adaptively estimate the importance of the semantic image feature since the variance of the illumination condition may introduce harmful interference information. Hence, we further provide the analysis of the weight map $\mathbf{w}$ for the semantic image feature in the following.

\noindent\textbf{Analysis of the weight map in the LI-Fusion layer.} In a real scene, the camera image is usually disturbed by the illumination, suffering from underexposure and overexposure. To verify the effectiveness of the weight map $\mathbf{w}$ in alleviating the interference information brought by the unsatisfying camera image, we simulate the real environment by changing the illumination of the camera image. For each image in the KITTI dataset, we simulate the illumination variance through the transformation $y=a*x+b$, where $x$ and $y$ denote the original and transformed RGB value for a pixel. $a$ and $b$ represent the coefficient and the offset, respectively. We randomly lighten up~(resp. darken) the camera images in the KITTI dataset by setting $a$ to 3~(resp. 0.3) and $b$ to 5. The quantitative results are presented in Table~\ref{tab:ablation on influence of the weight map}. For comparison, we remove the image stream and use our model based on only the LiDAR as the baseline, which yields a 3D mAP of 83.87\%. We also provide the results of simply concatenating the RGB and LiDAR coordinates in the input level~(denoted by SC), which leads to an obvious performance decreasement of 1.08\% and demonstrates that images under poor quality is harmful for the 3D detection task. Besides, our method without estimating the weight map $\mathbf{w}$ also results in a decreasement of 0.69\%. However, with the guidance of the weight map $\mathbf{w}$, our method yields an improvement of 0.65\% compared to the baseline. It means that introducing the weight map can adaptively select the beneficial features and ignore those harmful features.

\begin{minipage}{0.94\textwidth}
\hfill
\begin{minipage}{0.54\textwidth}
\makeatletter\def\@captype{table}\makeatother
\setlength{\belowcaptionskip}{4pt}%
\caption{Comparison between the results our LI-Fusion module with and without estimating the weight map $\mathbf{w}$.}
\label{tab:ablation on influence of the weight map}
\scriptsize
\resizebox{\columnwidth}{!}{
\begin{tabular}{|l|c|c|c|c|c|}
\hline
Method &Easy &Moderate &Hard &3D mAP & Gain \\
\hline\hline
only LiDAR &90.87 &81.15 &79.59  &83.87 &-  \\
SC &90.73 &79.93 &77.70 &82.79 &\textcolor{red}{$\downarrow$ 1.08} \\
Ours (without $\mathbf{w}$)    &91.52 &80.08 &77.95  &83.18 &\textcolor{red}{$\downarrow$ 0.69}  \\
Ours &\textbf{91.65} &\textbf{81.77} &\textbf{80.13}  &\textbf{84.52} &\textcolor{blue}{ $\uparrow$ 0.65}  \\
\hline
\end{tabular}
}
\end{minipage}
\hfill
\begin{minipage}{0.40\textwidth}
\centering
\makeatletter\def\@captype{table}\makeatother
\setlength{\belowcaptionskip}{5pt}%
\caption{The results of our approach on three benchmarks of the the KITTI validation set~(\textit{Cars}).}
\label{tab:results on the kitti val set}
\scriptsize
\resizebox{\columnwidth}{!}{
\begin{tabular}{|l|c|c|c|c|}
\hline
 Benchmark & Easy & Moderate & Hard & mAP\\
\hline\hline
	3D Detection    &92.28 &82.59 &80.14 &85.00 \\
	Bird’s Eye View &95.51 &88.76 & 88.36 &90.88 \\
	Orientation  &98.48 &91.74 &91.16  &93.79 \\
\hline
\end{tabular}
}
\end{minipage}
\end{minipage}

\begin{figure}[t]
\begin{center}
  \includegraphics[width=0.9\linewidth]{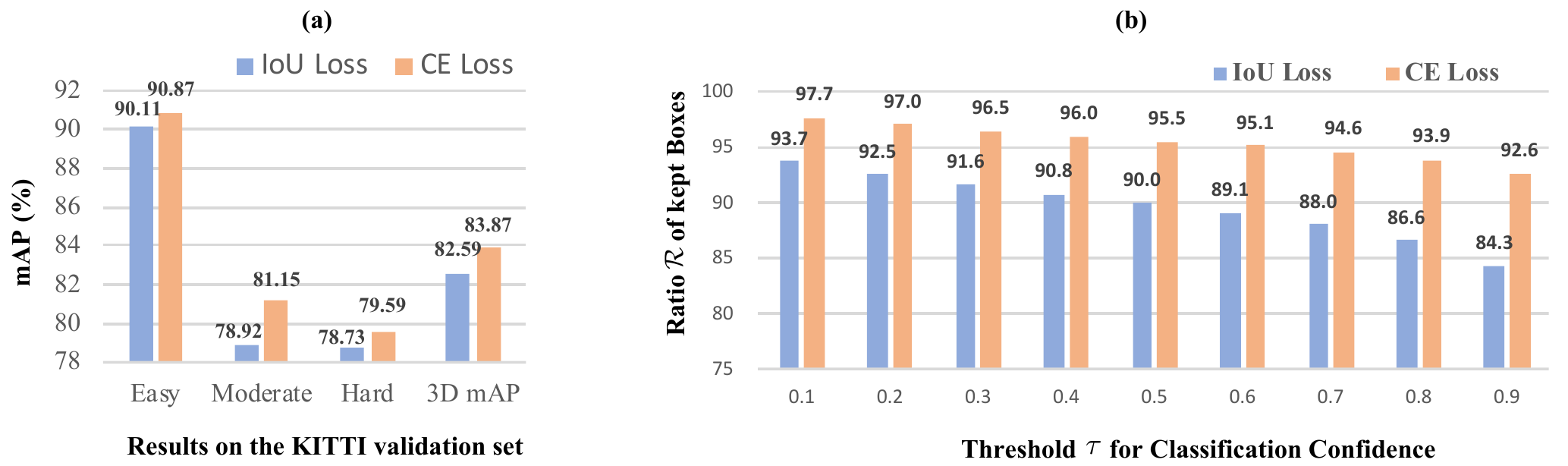}
\end{center}
\setlength{\abovecaptionskip}{2pt}
\caption{Illustration of the ratio of kept positive candidate boxes varying with different classification confidence threshold. The CE loss leads to significantly larger ratios than those of the IoU loss, suggesting its effectiveness in improving the consistency of localization and classification confidence.}
\label{fig:consistency}
\end{figure}

\noindent\textbf{Analysis of the CE loss.}~As shown in Table~\ref{tab:ablation study}, adding the CE loss yields a significant improvement of 3.93\% over the baseline. We further present a quantitative comparison with the IoU loss to verify the superiority of our CE loss in improving the 3D detection performance. As shown in Fig.~\ref{fig:consistency}(a), the CE loss leads to an improvement of 3D mAP of 1.28\% over the IoU loss, which indicates the benefits of ensuring the consistency of the classification and localization confidence in the 3D detection task.

To figure out how the consistency between these two confidences is improved, we give a thorough analysis of the CE loss. For the convenience of description, we denote predicted boxes possessing overlaps larger than a predefined IoU threshold $\tau$ as positive candidate boxes. Moreover, we adopt another threshold of $\upsilon$ to filter positive candidate boxes with smaller classification confidence. Hence, the consistency can be evaluated by the ratio of $\mathcal{R}$ of how many positive candidate boxes are kept, which can be written as follows:
\begin{equation}
\mathcal{R} = \frac{\mathcal{N}(\mathbf{b}|\mathbf{b} \in \mathcal{B} \  and \ \mathbf{c_{b}} > \upsilon)}{\mathcal{N}(\mathcal{B})},
\end{equation}
where $\mathcal{B}$ represents the set of positive candidate boxes. $\mathbf{c_{b}}$ denotes the classification confidence of the box $\mathbf{b}$. $\mathcal{N}(\cdot)$ calculates the number of boxes. It should be noted that all the boxes in $\mathcal{B}$ possess an overlap larger than $\tau$ with the corresponding ground truth box.

We provide evaluation results on two different settings, \textit{i.e.}, the model trained with IoU loss and that trained with CE loss. For each frame in the KITTI validation dataset, the model generates 64 boxes without NMS procedure employed. Then we get the positive candidate boxes by calculating the overlaps with the ground truth boxes. We set $\tau$ to 0.7 following the evaluation protocol of 3D detection metric. $\upsilon$ is varied from 0.1 to 0.9 to evaluate the consistency under different classification confidence thresholds. As is shown in Fig.~\ref{fig:consistency}(b), the model trained with CE loss demonstrates better consistency than that trained with IoU loss in all the different settings of classification confidence threshold $\upsilon$.


\subsection{Experiments on KITTI Dataset}
\label{sec:experiments on kitti}
 Table ~\ref{tab:results on kitti testing split} presents quantitative results on the KITTI test set. The proposed method outperforms multi-sensor based methods F-PointNet~\cite{qi2018frustum}, MV3D~\cite{chen2017multi}, AVOD-FPN~\cite{ku2018joint}, PC-CNN~\cite{8461232}, ContFuse~\cite{Liang2018ECCV}, and MMF~\cite{liang2019multi} by 10.37\%, 17.03\%, 7.71\%, 6.23\%, 9.85\% and 2.55\% in terms of 3D mAP. It should be noted that MMF~\cite{liang2019multi} exploits multiple auxiliary tasks~(\textit{e.g.}, 2D detection, ground estimation, and depth completion) to boost the 3D detection performance, which requires many extra annotations. These experiments consistently reveal the superiority of our method over the cascading approach~\cite{qi2018frustum}, as well as fusion approaches based on RoIs~\cite{chen2017multi,ku2018joint,8461232} and voxels~\cite{Liang2018ECCV,liang2019multi}.


\begin{table}[t]
	\scriptsize
	\centering
	\setlength{\belowcaptionskip}{2pt}
	\caption{Comparisons with state-of-the-art methods on the testing set of the KITTI dataset~(\textit{Cars}). L and I represent the LiDAR point cloud and the camera image.}
	\label{tab:results on kitti testing split}
	\resizebox{1\columnwidth}{!}{
		\begin{tabular}{|l|c|c|c|c|c|c|c|c|c|c|c|c|c|}
			\hline
			\multirow{2}{*}{Method} &\multirow{2}{*}{Modality} & \multicolumn{4}{c|}{3D Detection} &
			\multicolumn{4}{c|}{Bird's Eye View} &  \multicolumn{4}{c|}{Orientation}\\
			\cline{3-14}
			
			& & Easy & Moderate & Hard & 3D mAP & Easy & Moderate & Hard & BEV mAP & Easy & Moderate & Hard & Ori mAP\\
			\hline
			\hline
			SECOND~\cite{yan2018second}&L & 83.34 & 72.55 & 65.82 &73.90 & 89.39 & 83.77 & 78.59 &83.92 & 90.93 & 82.55 & 73.62 &82.37\\  
			PointPillars~\cite{lang2019pointpillars}&L & 82.58 & 74.31 & 68.99 &75.29 & 90.07 & 86.56 & 82.81 &86.48 & 93.84 & 90.70 & 87.47 &90.67\\ 
			TANet~\cite{liu2020tanet}&L & 84.39 & 75.94 & 68.82 &76.38 & 91.58 & 86.54 & 81.19 &86.44 & 93.52 & 90.11 & 84.61 &89.41\\ 
			PointRCNN~\cite{shi2019pointrcnn}&L  & 86.96 & 75.64 & 70.70 &77.77 & 92.13 & 87.39 & 82.72 &87.41 & 95.90 & 91.77 & 86.92 &91.53\\  
			Fast Point R-CNN~\cite{Chen2019fastpointrcnn}&L  & 85.29  & 77.40  &70.24 &77.64  & 90.87 & 87.84 & 80.52 &86.41 &- &- &- &- \\  
			\hline\hline
			F-PointNet~\cite{qi2018frustum}&L+I   & 82.19 & 69.79 & 60.59 &70.86 & 91.17 & 84.67 & 74.77  &83.54 &- &- &- &-\\  
			MV3D~\cite{chen2017multi}&L+I & 74.97 & 63.63 & 54.00 &64.20 & 86.62 & 78.93 & 69.80 &78.45 &- &- &- &-\\
			AVOD~\cite{ku2018joint}&L+I & 76.39  & 66.47 & 60.23 &67.70 & 89.75 & 84.95 & 78.32 &84.34 & 94.98 & 89.22 & 82.14 &88.78 \\ 
			AVOD-FPN~\cite{ku2018joint}&L+I & 83.07 & 71.76 & 65.73 &73.52 & 90.99 & 84.82 & 79.62 &85.14 &94.65 & 88.61 & 83.71 &88.99 \\  
			ContFuse~\cite{Liang2018ECCV}&L+I   &83.68 & 68.78 & 61.67 &71.38 & 94.07 & 85.35 & 75.88 &85.10 &- &- &- &- \\
			PC-CNN~\cite{8461232}&L+I  &85.57  &73.79 &65.65 &75.00 &91.19 &87.40 &79.35 &85.98 &- &- &- &-\\
			MMF~\cite{liang2019multi}&L+I  & 88.40 & 77.43 & 70.22 &78.68 & 93.67 & 88.21 & 81.99 &87.96 &- &- &- &- \\  
			\hline \hline
			Ours &L+I & \textbf{89.81} & \textbf{79.28} & \textbf{74.59} &\textbf{81.23} & \textbf{94.22} & \textbf{88.47} & \textbf{83.69} &\textbf{88.79} & \textbf{96.13} & \textbf{94.22} & \textbf{89.68} &\textbf{93.34}\\  	
			\hline
			
		\end{tabular}
	}
\end{table}

We also provide the quantitative results on the KITTI validation split in the Table~\ref{tab:results on the kitti val set} for the convenience of comparison with future work. Besides, we present the qualitative results on the KITTI validation dataset in the supplementary materials.

\subsection{Experiments on SUN-RGBD Dataset}
\label{sec:experiments on sun-rgbd}
We further conduct experiments on the SUN-RGBD dataset to verify the effectiveness of our approach in the indoor scenes. Table~\ref{tab:results on the sun-rgbd dataset.} demonstrates the results compared with the state-of-the-art methods. Our EPNet achieves superior detection performance, outperforming PointFusion~\cite{xu2017pointfusion} by 15.7\%, COG~\cite{ren2016three} by 12.2\%, F-PointNet~\cite{qi2018frustum} by 5.8\% and VoteNet~\cite{qi2019deep} by 2.1\% in terms of 3D mAP. The comparisons with multi-sensor based methods PointFusion~\cite{xu2017pointfusion} and F-PointNet~\cite{qi2018frustum} are especially valuable. Both of them first generate 2D bounding boxes from camera images using 2D detectors and then outputs the 3D boxes in a cascading manner. Specifically, F-PointNet utilizes only the LiDAR data to predict the 3D boxes. PointFusion combines global image features and points features in a concatenation fashion. Different from them, our method explicitly establishes the correspondence between point features and camera image features, thus providing finer and more discriminative representations. Besides, we provide the qualitative results on the SUN-RGBD dataset in the supplementary materials.

\begin{table}[t]
	\scriptsize
	\centering
	\setlength{\abovecaptionskip}{2pt}
	\setlength{\belowcaptionskip}{4pt}
	\caption{Quantitative comparisons with the state-of-the-art methods on the SUN-RGBD test set. P and I represent the point cloud and the camera image.}
	\label{tab:results on the sun-rgbd dataset.}
	\resizebox{1\columnwidth}{!}{
		\begin{tabular}{|l|c|c|c|c|c|c|c|c|c|c|c|c|}
			\hline
			Method & Modality& bathtub  & bed  & bookshelf & chair & desk & dresser & nightstand & sofa  &  table  & toilet  &3D mAP \\
			\hline
			\hline
			DSS~\cite{song2016deep} &P + I  & 44.2 & 78.8 & 11.9 & 61.2 & 20.5 & 6.4 & 15.4 & 53.5 & 50.3 & 78.9 & 42.1\\
			2d-driven~\cite{lahoud20172d} &P + I  & 43.5 & 64.5 & 31.4 & 48.3 & 27.9 & 25.9 & 41.9 & 50.4 & 37.0 & 80.4 & 45.1\\
			COG~\cite{ren2016three} &P + I  & 58.3  & 63.7 & 31.8 & 62.2 & \textbf{45.2} & 15.5 & 27.4 & 51.0 & \textbf{51.3} & 70.1 & 47.6 \\
			PointFusion~\cite{xu2017pointfusion} &P + I  & 37.3 & 68.6 & \textbf{37.7} & 55.1 & 17.2 & 24.0 & 32.3 & 53.8 & 31.0 & 83.8 & 44.1 \\
			F-PointNet~\cite{qi2018frustum} &P + I  & 43.3  & 81.1 & 33.3 & 64.2 & 24.7 & 32.0 & 58.1 & 61.1 & 51.1 & \textbf{90.9} & 54.0 \\
			VoteNet~\cite{qi2019deep} &P &74.4  &83.0 &28.8 &\textbf{75.3} &22.0 &29.8 &62.2 &64.0 &47.3 &90.1 & 57.7  \\
			\hline \hline
			Ours &P + I  & \textbf{75.4} & \textbf{85.2} & \textbf{35.4} & 75.0  & 26.1   & 31.3  &62.0  & \textbf{67.2} &\textbf{52.1}  & 88.2   & \textbf{59.8} \\
			
			\hline 
		\end{tabular}
	}
\end{table}

\section{Conclusion}

We have presented a new 3D object detector named EPNet, which consists of a two-stream RPN and a refinement network. The two-stream RPN reasons about different sensors~(\textit{i.e.}, LiDAR point cloud and camera image) jointly and enhances the point features with semantic image features effectively by using the proposed LI-Fusion module. Besides, we address the issue of inconsistency between the classification and localization confidence by the proposed CE loss, which explicitly guarantees the consistency between the localization and classification confidence. Extensive experiments have validated the effectiveness of the LI-Fusion module and the CE loss. In the future, we are going to explore how to enhance the image feature representation with depth information of the LiDAR point cloud instead, and its application in 2D detection tasks. 

\section*{Acknowledgement}
This work was supported by National Key R\&D Program of China (No.2018YFB\\1004600), Xiang Bai was supported by the National Program for Support of Top-notch Young Professionals and the Program for HUST Academic Frontier Youth Team 2017QYTD08.


\clearpage
%

\bibliographystyle{splncs04}
\bibliography{egbib}

\begin{thebibliography}{10}
\providecommand{\url}[1]{\texttt{#1}}
\providecommand{\urlprefix}{URL }
\providecommand{\doi}[1]{https://doi.org/#1}

\bibitem{chen2016monocular}
Chen, X., Kundu, K., Zhang, Z., Ma, H., Fidler, S., Urtasun, R.: Monocular 3d
  object detection for autonomous driving. In: Proc. of IEEE Intl. Conf. on
  Computer Vision and Pattern Recognition (2016)

\bibitem{chen20173d}
Chen, X., Kundu, K., Zhu, Y., Ma, H., Fidler, S., Urtasun, R.: 3d object
  proposals using stereo imagery for accurate object class detection. IEEE
  Trans. Pattern Anal. Mach. Intell.  \textbf{40}(5),  1259--1272 (2017)

\bibitem{chen2017multi}
Chen, X., Ma, H., Wan, J., Li, B., Xia, T.: Multi-view 3d object detection
  network for autonomous driving. In: Proc. of IEEE Intl. Conf. on Computer
  Vision and Pattern Recognition (2017)

\bibitem{Chen2019fastpointrcnn}
Chen, Y., Liu, S., Shen, X., Jia, J.: Fast point r-cnn. In: Porc. of IEEE Intl.
  Conf. on Computer Vision (2019)

\bibitem{8461232}
Du, X., Ang, M.H., Karaman, S., Rus, D.: A general pipeline for 3d detection of
  vehicles. In: 2018 IEEE International Conference on Robotics and Automation
  (ICRA). pp. 3194--3200 (May 2018). \doi{10.1109/ICRA.2018.8461232}

\bibitem{geigerwe}
Geiger, A., Lenz, P., Urtasun, R.: Are we ready for autonomous driving. In:
  Proc. of IEEE Intl. Conf. on Computer Vision and Pattern Recognition

\bibitem{girshick2015fast}
Girshick, R.: Fast r-cnn. In: Proceedings of the IEEE international conference
  on computer vision. pp. 1440--1448 (2015)

\bibitem{ioffe2015batch}
Ioffe, S., Szegedy, C.: Batch normalization: Accelerating deep network training
  by reducing internal covariate shift. In: Proc. of Intl. Conf. on Machine
  Learning (2015)

\bibitem{jiang2018acquisition}
Jiang, B., Luo, R., Mao, J., Xiao, T., Jiang, Y.: Acquisition of localization
  confidence for accurate object detection. In: Proc. of European Conference on
  Computer Vision (2018)

\bibitem{kingma2014adam}
Kingma, D.P., Ba, J.: Adam: A method for stochastic optimization. ICLR  (2014)

\bibitem{ku2018joint}
Ku, J., Mozifian, M., Lee, J., Harakeh, A., Waslander, S.L.: Joint 3d proposal
  generation and object detection from view aggregation. In: IROS. pp.~1--8.
  IEEE (2018)

\bibitem{ku2019monopsr}
Ku*, J., Pon*, A.D., Waslander, S.L.: Monocular 3d object detection leveraging
  accurate proposals and shape reconstruction. In: CVPR (2019)

\bibitem{lahoud20172d}
Lahoud, J., Ghanem, B.: 2d-driven 3d object detection in rgb-d images. In:
  Porc. of IEEE Intl. Conf. on Computer Vision (2017)

\bibitem{lang2019pointpillars}
Lang, A.H., Vora, S., Caesar, H., Zhou, L., Yang, J., Beijbom, O.:
  Pointpillars: Fast encoders for object detection from point clouds. In: Proc.
  of IEEE Intl. Conf. on Computer Vision and Pattern Recognition (2019)

\bibitem{li2019gs3d}
Li, B., Ouyang, W., Sheng, L., Zeng, X., Wang, X.: Gs3d: An efficient 3d object
  detection framework for autonomous driving. In: IEEE Conference on Computer
  Vision and Pattern Recognition (CVPR) (2019)

\bibitem{licvpr2019}
Li, P., Chen, X., Shen, S.: Stereo r-cnn based 3d object detection for
  autonomous driving. In: CVPR (2019)

\bibitem{liang2019multi}
Liang, M., Yang, B., Chen, Y., Hu, R., Urtasun, R.: Multi-task multi-sensor
  fusion for 3d object detection. In: Proc. of IEEE Intl. Conf. on Computer
  Vision and Pattern Recognition (2019)

\bibitem{Liang2018ECCV}
Liang, M., Yang, B., Wang, S., Urtasun, R.: Deep continuous fusion for
  multi-sensor 3d object detection. In: Proc. of European Conference on
  Computer Vision (2018)

\bibitem{lin2017focal}
Lin, T.Y., Goyal, P., Girshick, R., He, K., Doll{\'a}r, P.: Focal loss for
  dense object detection. In: Porc. of IEEE Intl. Conf. on Computer Vision
  (2017)

\bibitem{liu2019deep}
Liu, L., Lu, J., Xu, C., Tian, Q., Zhou, J.: Deep fitting degree scoring
  network for monocular 3d object detection. In: Proceedings of the IEEE
  Conference on Computer Vision and Pattern Recognition. pp. 1057--1066 (2019)

\bibitem{liu2020tanet}
Liu, Z., Zhao, X., Huang, T., Hu, R., Zhou, Y., Bai, X.: Tanet: Robust 3d
  object detection from point clouds with triple attention. In: AAAI. pp.
  11677--11684 (2020)

\bibitem{luo2018fast}
Luo, W., Yang, B., Urtasun, R.: Fast and furious: Real time end-to-end 3d
  detection, tracking and motion forecasting with a single convolutional net.
  In: Proc. of IEEE Intl. Conf. on Computer Vision and Pattern Recognition
  (2018)

\bibitem{ma2019accurate}
Ma, X., Wang, Z., Li, H., Zhang, P., Ouyang, W., Fan, X.: Accurate monocular
  object detection via color- embedded 3d reconstruction for autonomous
  driving. In: Proceedings of the IEEE international Conference on Computer
  Vision (ICCV) (2019)

\bibitem{lasernet}
Meyer, G.P., Laddha, A., Kee, E., Vallespi-Gonzalez, C., Wellington, C.K.:
  {LaserNet}: An efficient probabilistic 3{D} object detector for autonomous
  driving. In: Proceedings of the IEEE Conference on Computer Vision and
  Pattern Recognition (CVPR) (2019)

\bibitem{mousavian20173d}
Mousavian, A., Anguelov, D., Flynn, J., Kosecka, J.: 3d bounding box estimation
  using deep learning and geometry. In: Proc. of IEEE Intl. Conf. on Computer
  Vision and Pattern Recognition (2017)

\bibitem{qi2019deep}
Qi, C.R., Litany, O., He, K., Guibas, L.J.: Deep hough voting for 3d object
  detection in point clouds. Porc. of IEEE Intl. Conf. on Computer Vision
  (2019)

\bibitem{qi2018frustum}
Qi, C.R., Liu, W., Wu, C., Su, H., Guibas, L.J.: Frustum pointnets for 3d
  object detection from rgb-d data. In: Proc. of IEEE Intl. Conf. on Computer
  Vision and Pattern Recognition (2018)

\bibitem{qi2017pointnet++}
Qi, C.R., Yi, L., Su, H., Guibas, L.J.: Pointnet++: Deep hierarchical feature
  learning on point sets in a metric space. In: Advances in neural information
  processing systems. pp. 5099--5108 (2017)

\bibitem{qin2019monogrnet}
Qin, Z., Wang, J., Lu, Y.: Monogrnet: A geometric reasoning network for 3d
  object localization. The Thirty-Third AAAI Conference on Artificial
  Intelligence (AAAI-19)  (2019)

\bibitem{ren2016three}
Ren, Z., Sudderth, E.B.: Three-dimensional object detection and layout
  prediction using clouds of oriented gradients. In: Proc. of IEEE Intl. Conf.
  on Computer Vision and Pattern Recognition (2016)

\bibitem{shi2019pointrcnn}
Shi, S., Wang, X., Li, H.: Pointrcnn: 3d object proposal generation and
  detection from point cloud. In: Proc. of IEEE Intl. Conf. on Computer Vision
  and Pattern Recognition (2019)

\bibitem{simonelli2019disentangling}
Simonelli, A., Bul{\`o}, S.R.R., Porzi, L., L{\'o}pez-Antequera, M.,
  Kontschieder, P.: Disentangling monocular 3d object detection. arXiv preprint
  arXiv:1905.12365  (2019)

\bibitem{song2015sun}
Song, S., Lichtenberg, S.P., Xiao, J.: Sun rgb-d: A rgb-d scene understanding
  benchmark suite. In: Proc. of IEEE Intl. Conf. on Computer Vision and Pattern
  Recognition (2015)

\bibitem{song2016deep}
Song, S., Xiao, J.: Deep sliding shapes for amodal 3d object detection in rgb-d
  images. In: Proc. of IEEE Intl. Conf. on Computer Vision and Pattern
  Recognition (2016)

\bibitem{wangcvpr2019}
Wang, Y., Chao, W.L., Garg, D., Hariharan, B., Campbell, M., Weinberger, K.:
  Pseudo-lidar from visual depth estimation: Bridging the gap in 3d object
  detection for autonomous driving. In: CVPR (2019)

\bibitem{xu2018multi}
Xu, B., Chen, Z.: Multi-level fusion based 3d object detection from monocular
  images. In: Proc. of IEEE Intl. Conf. on Computer Vision and Pattern
  Recognition (2018)

\bibitem{xu2017pointfusion}
Xu, D., Anguelov, D., Jain, A.: Pointfusion: Deep sensor fusion for 3d bounding
  box estimation. In: Proc. of IEEE Intl. Conf. on Computer Vision and Pattern
  Recognition (2018)

\bibitem{yan2018second}
Yan, Y., Mao, Y., Li, B.: Second: Sparsely embedded convolutional detection.
  Sensors  \textbf{18}(10), ~3337 (2018)

\bibitem{yang2018pixor}
Yang, B., Luo, W., Urtasun, R.: Pixor: Real-time 3d object detection from point
  clouds. In: Proc. of IEEE Intl. Conf. on Computer Vision and Pattern
  Recognition (2018)

\bibitem{std2019yang}
Yang, Z., Sun, Y., Liu, S., Shen, X., Jia, J.: {STD:} sparse-to-dense 3d object
  detector for point cloud. ICCV  (2019), \url{http://arxiv.org/abs/1907.10471}

\bibitem{yu2016unitbox}
Yu, J., Jiang, Y., Wang, Z., Cao, Z., Huang, T.: Unitbox: An advanced object
  detection network. In: Proceedings of the 24th ACM international conference
  on Multimedia (2016)

\bibitem{zhao20193d}
Zhao, X., Liu, Z., Hu, R., Huang, K.: 3d object detection using scale invariant
  and feature reweighting networks. In: Proceedings of the AAAI Conference on
  Artificial Intelligence. vol.~33, pp. 9267--9274 (2019)

\bibitem{zhou2018voxelnet}
Zhou, Y., Tuzel, O.: Voxelnet: End-to-end learning for point cloud based 3d
  object detection. In: Proc. of IEEE Intl. Conf. on Computer Vision and
  Pattern Recognition (2018)

\end{thebibliography}

\clearpage

\begin{appendix}

\section{More Qualitative Results}
In this section, we first present more qualitative results on the KITTI and the SUN-RGBD dataset. Then we provide several qualitative analyses of the effect of the LI-Fusion module in the 3D object detection task.

\subsection{KITTI Dataset}
\begin{figure*}[h]
\begin{center}
  \includegraphics[width=0.90\linewidth]{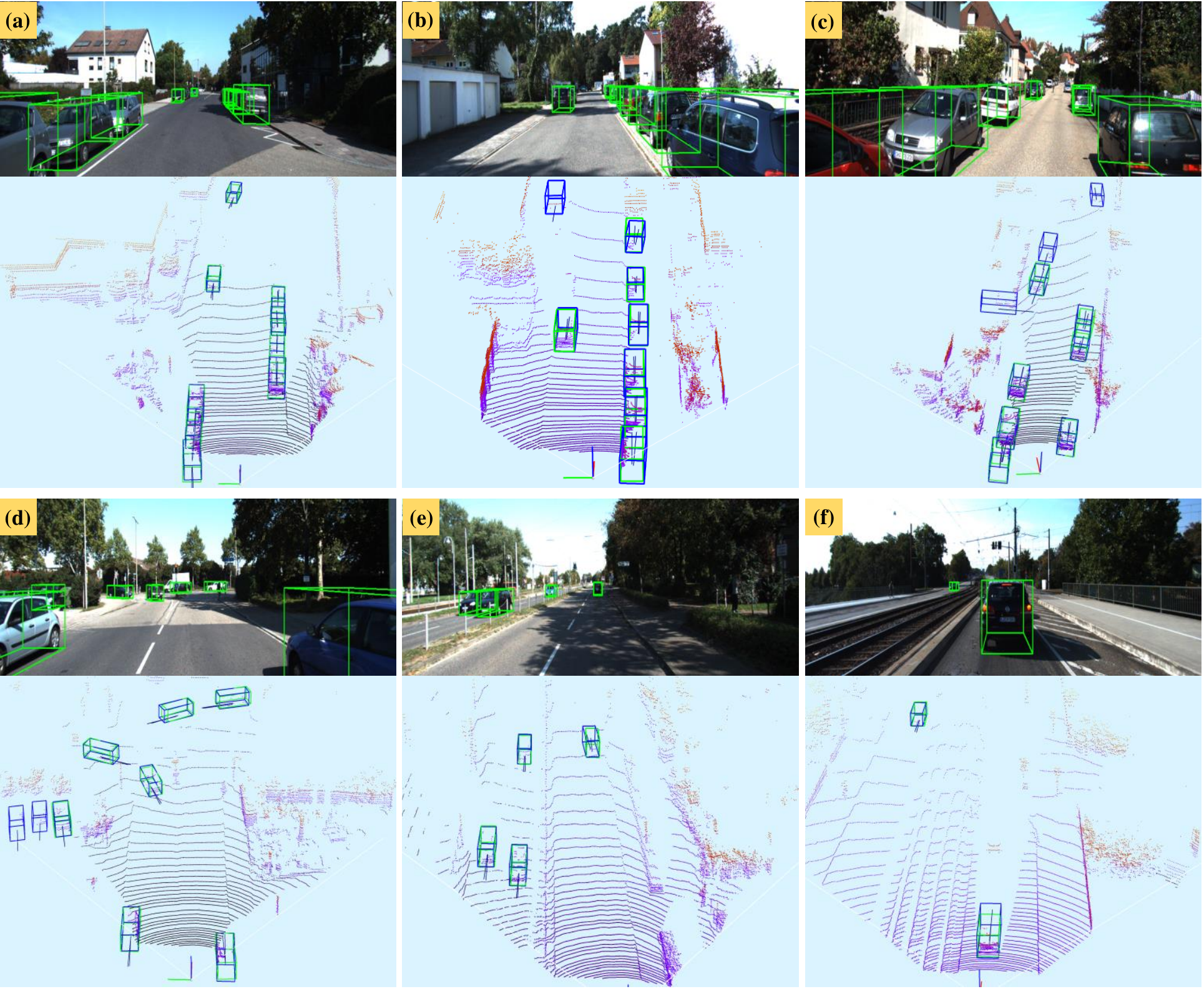}
\end{center}
\caption{Qualitative results of our approach on the KITTI validation set. For each pair, the first and the second row show the camera image and the representative view of LiDAR point cloud. The ground truth and detected boxes are highlighted with green and blue boxes, respectively.}
\label{fig:vis of kitti}
\end{figure*}

Fig.~\ref{fig:vis of kitti} illustrates the qualitative results on the KITTI validation set. Our method can detect the objects in the 3D scene accurately. On the one hand, our method produces precise boxes even under the extremely challenging cases where multiple cars are crowded together, as is shown in Fig.~\ref{fig:vis of kitti}~(a), (b), and (c). On the other hand, our approach captures the cars far away well~(\textit{e.g.} Fig.~\ref{fig:vis of kitti}~(d), (e), and (f)), although these objects are usually difficult to be recognized in the camera image and suffer from the sparsity of the point cloud. All these challenging cases persuasively demonstrate the effectiveness of our method.


\subsection{SUN-RGBD Dataset}

\begin{figure*}[h]
\begin{center}
  \includegraphics[width=0.92\linewidth]{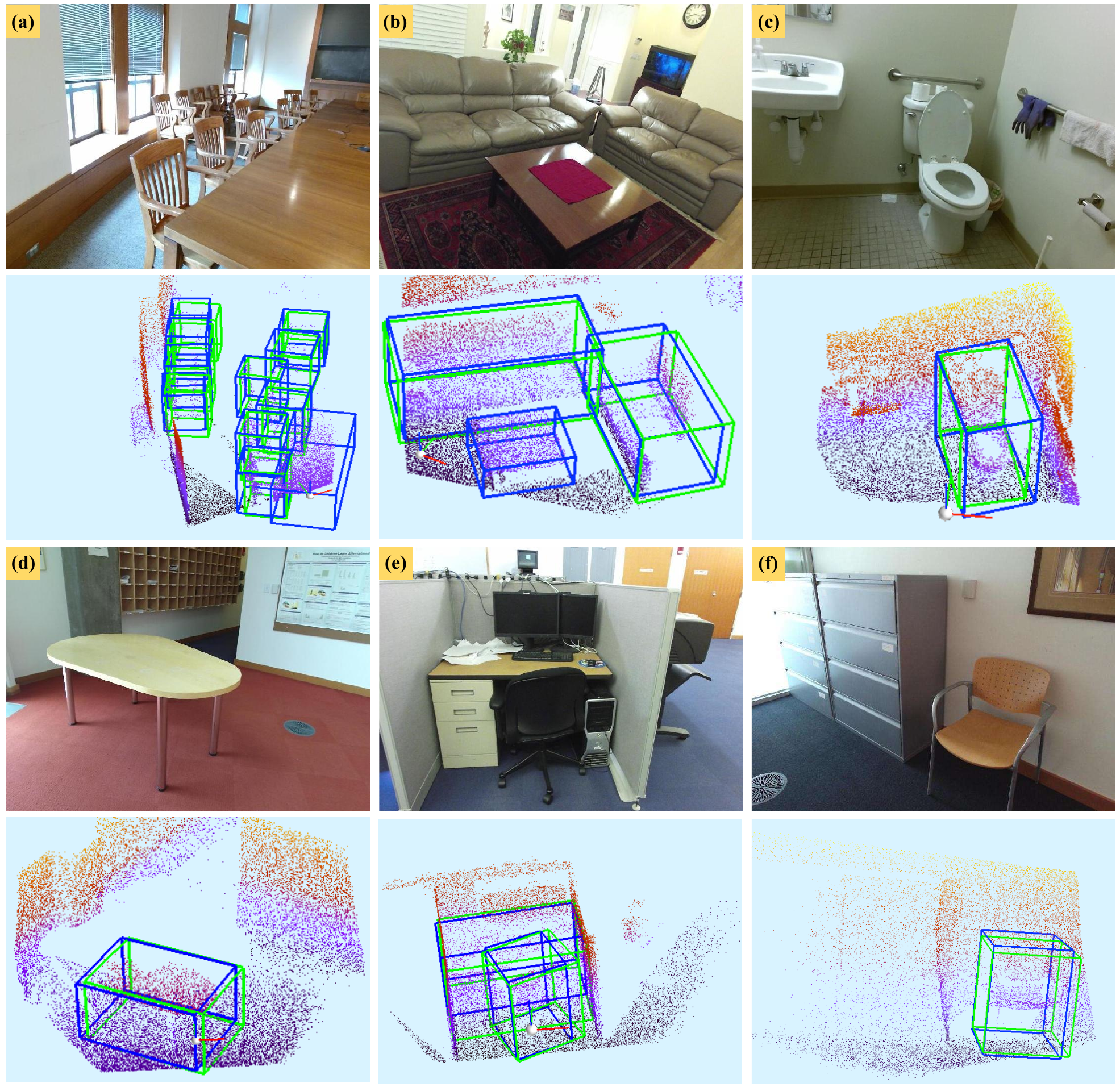}
\end{center}
\caption{Qualitative results of our approach on the SUN-RGBD test set. For each pair, the first and the second row show the camera image and the representative view of LiDAR point cloud. The ground truth and detected boxes are highlighted with green and blue boxes, respectively. }
\label{fig:vis of sun-rgbd}
\end{figure*}

We present the qualitative results on the SUN-RGBD test set in Fig.~\ref{fig:vis of sun-rgbd}. Different from the KITTI dataset, SUN-RGBD is an indoor dataset which contains objects of many categories and various scales. As is shown in Fig.~\ref{fig:vis of sun-rgbd}, our method and can accurately detect multiple kinds of objects with significant scale variations, including large objects~(\textit{e.g.}, bed, sofa) and small objects~(\textit{e.g.}, dresser, chair). Predicting the bounding boxes of objects in a crowded area is especially challenging. For example, Fig.~\ref{fig:vis of sun-rgbd}~(a) are crowded with lots of chairs and increase the difficulty for detection significantly. Even under this challenging case, our method still outputs precise bounding boxes, demonstrating the robustness of our method for crowded objects.


\subsection{Analysis on the LI-Fusion Module}
As mentioned in the main manuscript, the LI-Fusion module can combine the abundant semantic information~(\textit{e.g.}, color) in the camera image and the geometric information encoded in the LiDAR point cloud. In this section, we provide a qualitative analysis on the effect of the LI-Fusion module. 

\begin{figure*}
\begin{center}
  \includegraphics[width=0.95\linewidth]{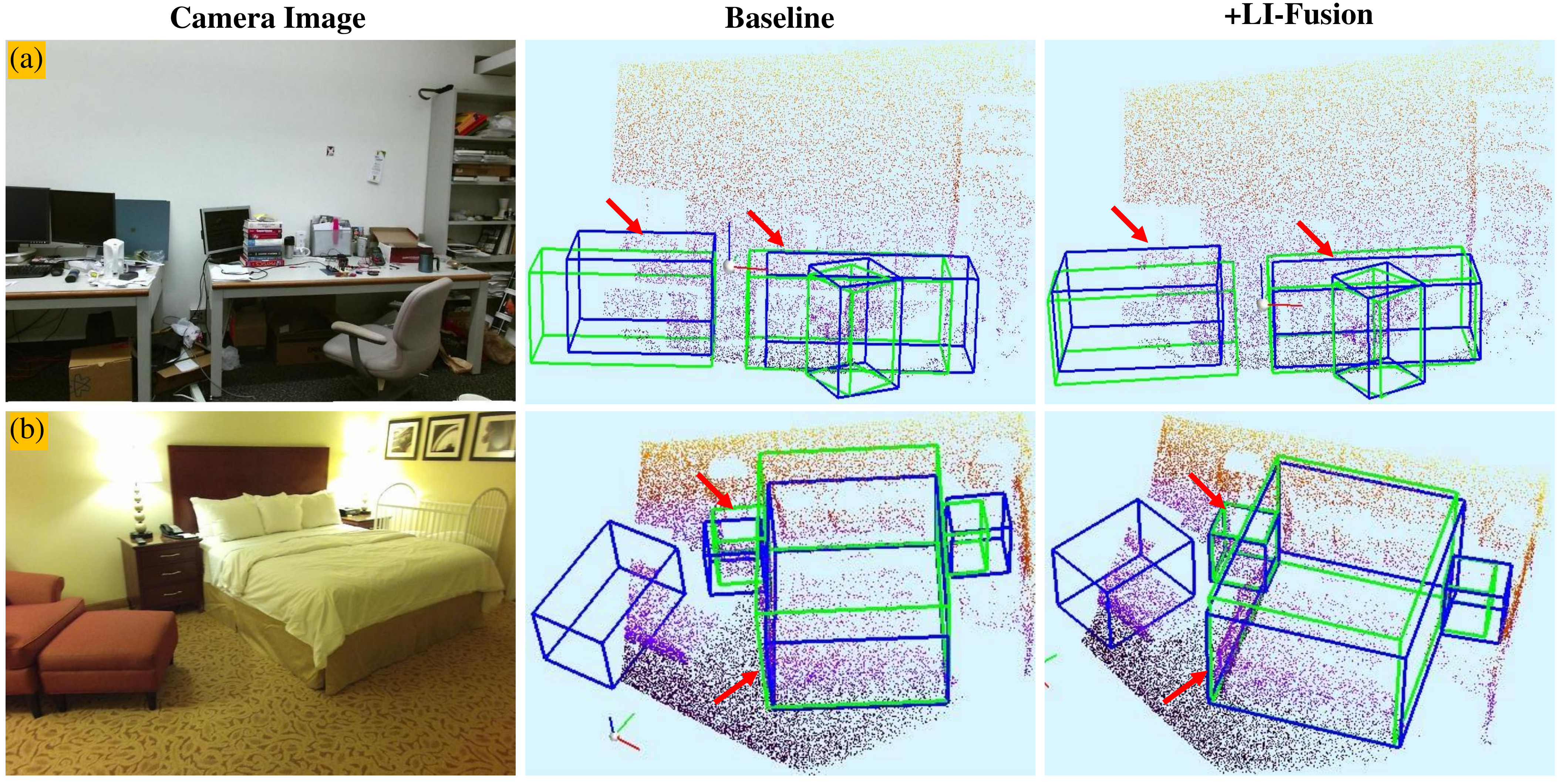}
\end{center}
\caption{Qualitative analysis of the effect of our LI-Fusion module on the SUN-RGBD test set. The LI-Fusion module effectively exploits the semantic image information, which is important for generating more bounding boxes .}
\label{fig:comp of LI-Fusion}
\end{figure*}

We remove the LI-Fusion module from our EPNet and compare its results with those of EPNet. As shown in Fig.~\ref{fig:comp of LI-Fusion}, the LI-Fusion module leads to more precise bounding boxes. The rationale behind is that the edge information and the color information embedded in the camera image help differentiate an object from its neighboring environment, for example, the desks in Fig.~\ref{fig:comp of LI-Fusion}(a), as well as the bed and the dresser in Fig.~\ref{fig:comp of LI-Fusion}(b). These results consistently verify the effectiveness of our LI-Fusion module in exploiting the semantic image information and LiDAR point cloud information for improving the 3D detection task. 


\subsection{Visualization for Images with Varying Illumination}
\begin{figure*}[t]
\begin{center}
  \includegraphics[width=0.95\linewidth]{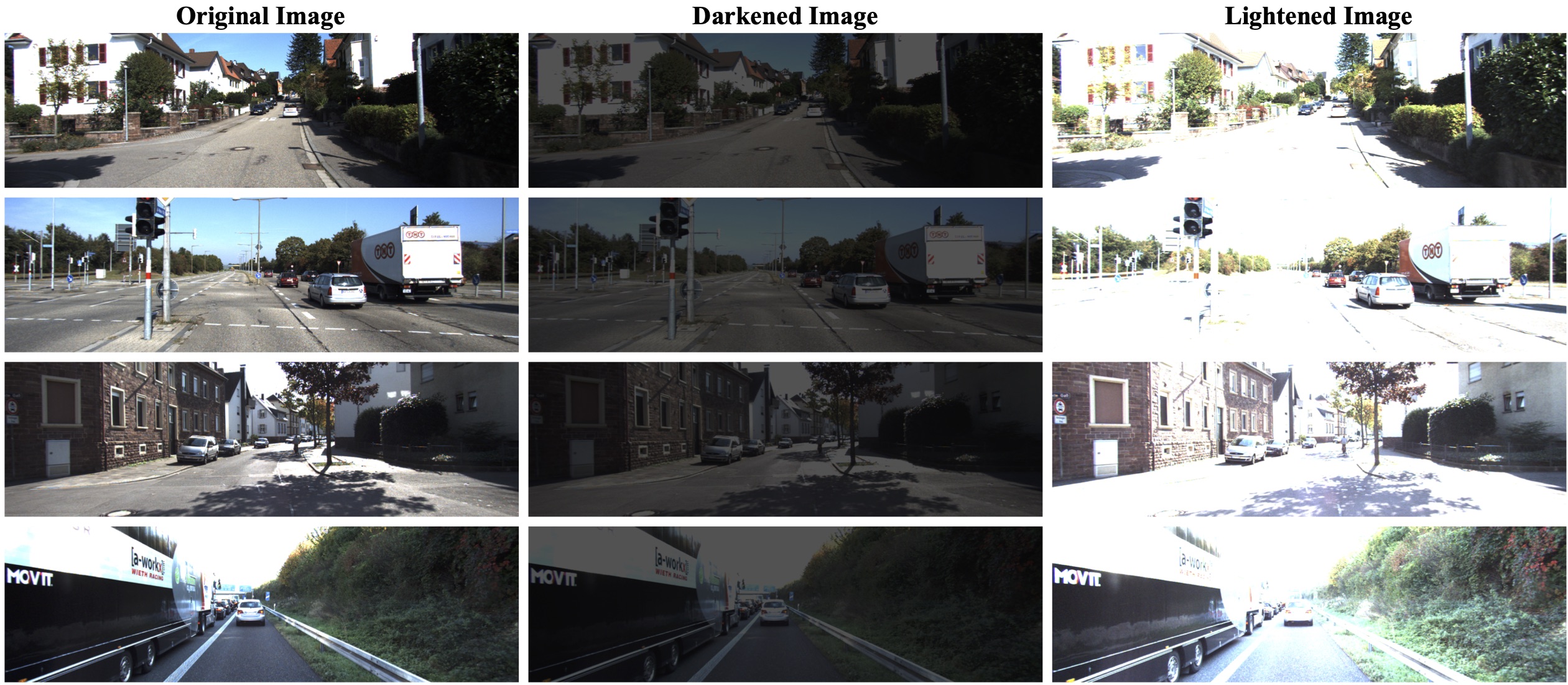}
\end{center}
\caption{Visualization of the darkened images and lightened images generated by the illumination transformation to simulate the underexposure and overexposure cases in the real scenes.}
\label{fig:vis of illumination}
\end{figure*}

In the main manuscript, we simulate the real environment by varying the illumination condition through a transformation function and verify the effectiveness of our LI-Fusion module. In Fig.~\ref{fig:vis of illumination}, we provide some examples generated by the transformation function. It can be seen that the darkened images and the lightened images can well simulate the underexposure and the overexposure cases in the real scenes. Even under such severe illumination conditions, where the camera images bring much interfering information, our LI-Fusion can still effectively enhance the point features and lead to improved detection performance as shown in the main manuscript. It demonstrates the superiority of the LI-Fusion module in adaptively selecting the beneficial features and suppressing the harmful features.

\end{appendix}

\end{document}